\documentclass[letterpaper, 10 pt, conference]{ieeeconf}

\IEEEoverridecommandlockouts                              

\usepackage{amssymb}
\usepackage{hyperref}
\usepackage{amsmath}
\usepackage{algorithm}
\usepackage{algorithmicx}
\usepackage{algpseudocode}
\usepackage{graphicx}
\usepackage{pgfplots}
\usepackage{tikz}
\usepackage{booktabs}

\newcommand{\mypara}[1]{\vspace{0.7em}\noindent\textbf{#1}}

\usepackage[caption=false, font=footnotesize]{subfig}

\pgfplotsset{every axis/.append style={
                    label style={font=\small},
                    tick label style={font=\small}  
                    }}

\title{\LARGE \bf
Getting to Know One Another: Calibrating Intent, Capabilities and Trust for Human-Robot Collaboration
}

\author{Joshua Lee, Jeffrey Fong, Bing Cai Kok, and Harold Soh\\Dept. of Computer Science, National University of Singapore\\{\small\texttt{joshua\_lks@u.nus.edu, \{jfong, kokbc, harold\}@comp.nus.edu.sg}}}

\begin{document}

\maketitle
\thispagestyle{empty}
\pagestyle{empty}

\begin{abstract}
Common experience suggests that agents who know each other well are better able to work together. In this work, we address the problem of calibrating intention and capabilities in human-robot collaboration. In particular, we focus on  scenarios where the robot is attempting to assist a human who is unable to directly communicate her intent. Moreover, both agents may have differing capabilities that are unknown to one another. We adopt a decision-theoretic approach and propose the TICC-POMDP for modeling this setting, with an associated online solver. Experiments show our approach leads to better team performance both in simulation and in a  real-world study with human subjects.
\end{abstract}
\section{INTRODUCTION}
As robots are increasingly deployed into our homes and workplaces, it is crucial that human users trust their robot collaborators appropriately. In particular, it is important to mitigate under-trust and over-trust in robots, which can lead to unsatisfactory outcomes~\cite{chen2020trust,robinette2016overtrust}. Building upon prior work that shows human trust in robots is dependent on the robot's intention (or policy) and  capabilities~\cite{xie2019robot,hancock2011meta}, we focus on \emph{calibrating} these two constructs for trust-based human-robot collaboration. 

We consider assistive scenarios where the robot is helping the human to accomplish a particular goal, but is unaware of the human's intent. As such, the robot has to learn the goal through interaction. Unlike a majority of existing work, we address the case where the human and robot have asymmetric capabilities that are unknown to one another. This setting is important as non-expert users may be unaware of the robot's programming and physical capabilities. Poorly calibrated capability models can damage human users' perception of the robot's trustworthiness~\cite{lee2010gracefully}, which in turn affects their willingness to cooperate with it. Likewise, humans have differing proficiency at tasks, and failure to recognize capability differences may lead to incorrect inferences about a person's goals. In short, failure to acknowledge each other's capabilities can impede collective performance. 

\begin{figure}
\centering
\includegraphics[width=0.85\linewidth]{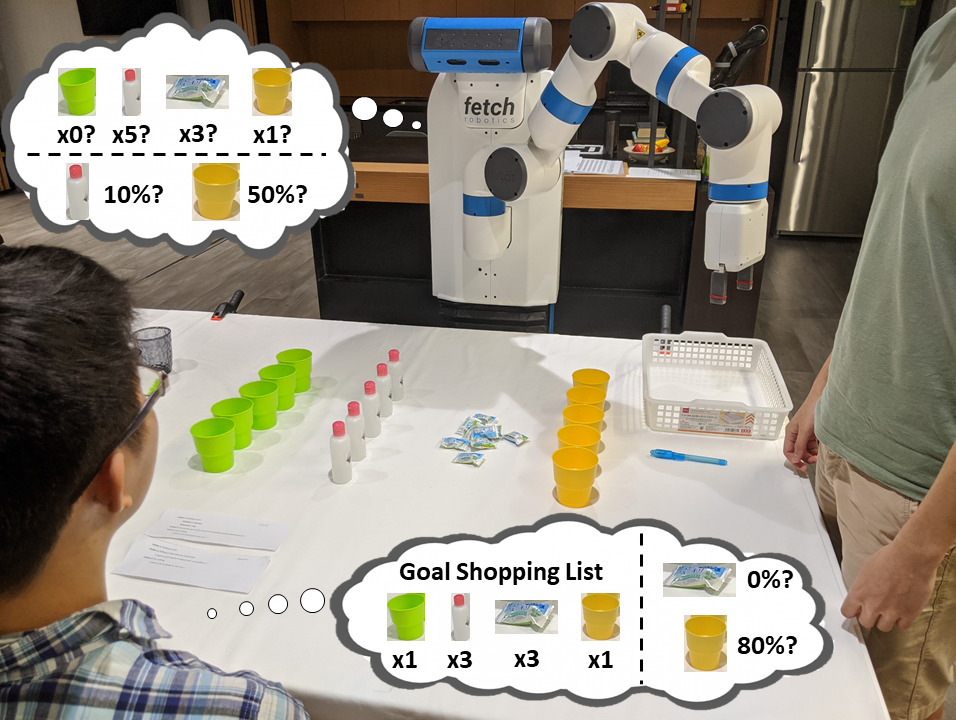}
\caption{Calibration experiment with human subjects. The human and the Fetch robot have to work together to complete a shopping task. However, only the human knows the goal shopping list, and is unable to communicate it directly to the robot. Moreover, both agents have imperfect capabilities and are unable to pick up certain items. Using our TICC-POMDP (and associated TICC-MCP solver), the agents undergo intent-capability calibration and update their beliefs of each others' capabilities over time. Experiments show that our approach leads to more accurate beliefs over intention and capabilities, and in-turn, higher task rewards and trust.}
\label{fig:human_exp_setup}
\vspace{-2mm}
\end{figure}

We undertake a decision theoretic approach to the problem and contribute the Trust-Intent-Capability-Calibration POMDP (TICC-POMDP). The key idea underlying our approach is a structured transition, which enables us to incorporate both robot and human capability models as parameters in an expanded state-space model. As estimations of capability are likely to persist over multiple rounds of interactions (and even tasks), we encourage calibration via an additional calibration reward that incentivizes the robot to engage in teaching behaviors that inform the human about its (in)capabilities. To solve the TICC-POMDP, we propose an online planning Monte-Carlo method, the TICC-MCP, which scales to large state-spaces. Code and an appendix are available at \href{https://github.com/clear-nus/TICC-MCP}{\small\texttt{https://github.com/clear-nus/TICC-MCP}}.

We present both simulation and real-world experiments with human subjects ($n=28$). Our simulation experiments focused on the algorithmic properties of the TICC-MCP and we find our method outperforms the standard POMCP (which lacks capability models). Our human subject experiments (Fig \ref{fig:human_exp_setup}) show that our approach earned higher rewards with actual humans and we find  evidence that suggests our method induced higher levels of trust in the robot.

To summarize, our key contributions are:
\begin{itemize}
	\item A novel decision-theoretic formulation for calibrating  intent and capability in human-robot collaboration;
	\item The TICC-MCP online planning algorithm for (approximately) solving the TICC-POMDP;
	\item Simulation and real-world human subject experimental findings showing that intent and capability calibration is beneficial over repeated interactions. 
\end{itemize}
\section{Background}

In this section, we briefly cover the necessary background material; we first give an overview of sequential decision-making frameworks, specifically the Markov decision process (MDP) and its generalization, the partially-observable MDP (POMDP) and Bayes Adaptive POMDP (BA-POMDP). Then, we discuss inverse reinforcement learning, and trust in human-robot collaboration. 

\subsection{Decision-Making Frameworks}

\mypara{Markov Decision Processes.} A MDP models a sequential decision-making problem as a tuple $\left< S,A,T,R,\gamma\right>$, where $S$ is the set of environmental states, $A$ is the set of actions available to the agent, and the transition function $T(s_t,a_t,s_{t+1})$ models the probability of transitioning to state $s_{t+1} \in S$ after taking action $a_t \in A$ in state $s_{t}$ . At each state, the agent receives a reward $R(s)$\footnote{The reward may also depend on the next state $s_{t+1}$ and action $a_t$.} and the agent's goal is to obtain the optimal policy $\pi^*: S \rightarrow A$ that maximizes the expected return:  
$\pi^* = \arg\max_\pi \mathbb{E}\left[ \sum_t \gamma^t R(s_t)\right]$
where  $\gamma\in[0,1]$ is a discount factor and the expectation is over the trajectories generated by following $\pi$.

\mypara{Partially-Observable MDPs.} In contrast to MDPs where states are fully-observable, POMDPs model sequential decision-making \emph{under uncertainty} where agents can only make observations that arise from the underlying state. We augment the MDP with two additional elements: the set of observations $Z$ and the observation function $O(s_{t+1},a_t,z_t) = p(z_t | s_{t+1}, a_t)$, which specifies the probability of observing $z_t \in Z$ after having taken action $a_t \in A$ and arriving in state $s_{t+1} \in S$.  Because the agent cannot observe the state directly, it has to rely on its interaction history $h_t= \left<a_o,z_1,… ,a_{t-1},z_t\right>$ to infer a belief $b(s)$ over states. The policy $\pi(b)$ thus maps beliefs onto actions. As in the MDP case, the agent aims to maximize its expected return given an initial belief $b_0$:
\begin{align}
    \pi^* = \arg\max_\pi V^\pi(b_0) =  \arg\max_\pi \mathbb{E}\left[ \sum_t \gamma^t R(s_t) \,\Big|\, b_0 \right]
\end{align}
where $V^\pi(b_0)$ is the value function and the expectation is over trajectories obtained by following $\pi$ starting from belief $b_0$. In general, solving POMDPs is intractable but efficient and scalable methods exist for approximate planning ~\cite{kurniawati2008sarsop, somani2013despot,karkus2017qmdp,pineau2003point}.

\mypara{Bayes-Adaptive POMDP.}
The environment model is often not perfectly known to the agent. The Bayes-Adaptive POMDP (BA-POMDP) \cite{ross2008bayes} is an example of the Bayesian Reinforcement Learning approach for learning POMDP transition and observation models. The BA-POMDP uses Dirichlet distributions to represent uncertainty over the transition and observation models; the agent maintains a vector $\chi$ with the experience counts of $\left<s,a,s',z\right>$. The agent is uncertain about the true count vector and this uncertainty is represented as hidden state in POMDP, i.e., the state space is augmented with the Dirichlet parameters. 

As BA-POMDPs have infinite state space, Ross \emph{et al.}~\cite{ross2008bayes} proposed a reduction to a finite model, and developed an online lookahead planner. Extending this work, Katt \emph{et al.}~\cite{katt2017learning} developed the BA-POMCP algorithm, which uses POMCP~\cite{silver2010monte} to solve BA-POMDPs more efficiently via sample-based online planning. Our collaborative model TICC-POMDP is a BA-POMDP with specific structure, and we propose a variant of BA-POMCP algorithm.

\subsection{Inferring Reward Functions}

In this paper, we cast intent inference as problem of inverse reinforcement learning (IRL), i.e., recovering the reward function of the agent given its policy or observed behavior~\cite{arora2018survey}. More formally, let ${\cal M}_{\backslash R}$ be an MDP without the reward function $R$ of an expert. IRL seeks a reward function $\hat{R}_E$ that best describes a set of observed expert demonstrations $D = \{\tau_i\}_i^N$. The approach underlying many IRL methods (e.g.,~\cite{russell1998learning,ng2000algorithms,abbeel2004apprenticeship}) is to update (or optimize) a parametric reward function $R(s; \theta)$ until it is consistent with the observed behavior, under constraints or heuristics to address ambiguity~\cite{arora2018survey,ng2000algorithms,ziebart2008maximum}.

One potential drawback of the above methods is the underlying assumption that the demonstrator acts to maximize expected returns. However,  expert demonstrations may be suboptimal due to cognitive biases in planning~\cite{bergen2010learning,evans2015learning,evans2016learning}, risk sensitivity~\cite{majumdar2017risk}, imperfect beliefs about real world dynamics~\cite{reddy2018you} and pedagogy behavior~\cite{hadfield2016cooperative}. Here, we consider situations where the expert demonstrations are ``suboptimal'' in that the robot and human may have incorrect models of each other's capabilities.

\subsection{Trust in Human Robot Collaboration}

Trust is important in shaping how humans interact with one another~\cite{balliet2013trust}. As with humans, the issue of human trust in automation has also been identified as a key determinant to successful human-robot collaboration~\cite{hancock2011meta, chen2020trust}. For instance, trust in automation has been shown to increase performance in human-robot teams~\cite{you2018trusting}, affect reliance on automation~\cite{dzindolet2003role} and determine whether or not humans will choose to retaliate against an adversarial robot~\cite{sebo2019don}. 

There exist various definitions of trust, ranging from a belief~\cite{lee2004trust} to an operator in a logic~\cite{huang2019reasoning}. 
We adopt a formal definition of trust in the human-robot collaboration literature: trust is a latent variable that summarizes a human's interactions with the robot and determines whether or not one would be willing to rely on the autonomous robot~\cite{chen2018planning, soh2020multi}.

Through this formalization, trust in automation can also be exploited to further improve both planning~\cite{chen2018planning} and predictive algorithms that can capture how human trust in a robot transfers across multiple tasks \cite{soh2018transfer}. Most pertinently, the explicit incorporation of human factors, such as the human's preference for explainability~\cite{wang2016trust}, into the design of autonomous agents could implicitly foster greater trust in autonomous robotic agents. In line with this, we explore the hypothesis that calibrated human and robot capability models can encourage humans to develop trust in robots.
\section{Calibrating Intent and Capabilities with Ticc-Pomdp}
\label{sec:problem}

In this section, we describe the TICC-POMDP for calibrating both intention and capabilities. We consider planning from the \emph{perspective of the robot} and at a high level, our formulation is a POMDP augmented with human and robot capability models. Specifically, we model the robot's belief over the human's capabilities and the human's belief over the robot's capabilities. These models enter into a \emph{structured} {transition function}. Overall, the TICC-POMDP can be seen as a BA-POMDP with latent reward and capability parameters, and a reward calibration component. 
 
 The following elaborates upon the above; we first describe the key ideas underlying our approach, and then apply them to the human-robot collaboration setting. 

\subsection{Action Success \& Failures and Factorized Transitions}

In our model, actions can either succeed or fail; for example, an agent may accidentally drop an object when trying to pick it up; the agent did not \emph{intend} to drop the item, but may do so due to its (lack of) dexterity at handling items, i.e., its \emph{capability}. In other words, we consider the capability of agents to be the probability that actions taken are successful. In typical POMDPs, action success/failure is typically \emph{implicit} within the model's transition probabilities. The key idea in our approach is to make these ``action outcomes'' \emph{explicit}. For expositional simplicity, we discuss binary success/failure action outcomes but our formulation generalizes to different levels of success/failure.  

More precisely, for every action $a$, we define a successful action $a^\text{S}$ and a failed action $a^\text{F}$. We can then model state transitions as a marginalization of a factorized transition probability over \emph{action outcomes} $\tilde{a}$:
\begin{align}
	P(s_{t+1}|s_t, a_t) & = \sum_{\tilde{a}_t \in \{a_t^\text{S}, a_t^\text{F}\} } P(s_{t+1}| s_t, \tilde{a}_t, a_t) P(\tilde{a}_t|s_t, a_t) \nonumber \\
	& = \sum_{\tilde{a}_t \in \{a_t^\text{S}, a_t^\text{F}\} } P(s_{t+1}| s_t, \tilde{a}_t) P(\tilde{a}_t| a_t)
	\label{eqn:Tprob}
\end{align}
where we have assumed that (i) given the action outcome $\tilde{a}_t$, the next state is independent of the action $a_t$, and (ii) given the action $a_t$, the action outcome $\tilde{a}_t$ is independent of the state $s_t$. The latter conditional independence assumption is not strictly necessary but reduces model complexity. 

Intuitively, the structured transition comprises two parts:
\begin{align}
	\underbrace{P(s_{t+1}| s_t, \tilde{a}_t)}_\text{World Physics} \underbrace{P(\tilde{a}_t|a_t)}_\text{Agent Capability}
\end{align}
Decomposing the transition probability in this manner has benefits; it represents the world physics separately from the agent's capabilities, which can ease data requirements if the transition model is learnt from data---particularly if one part (e.g., the world physics) is known. Moreover, different agent models can be substituted into the POMDP without having to re-learn or re-model the entire transition dynamics. Likewise, the same agent model can be applied to different environments (with the same action set).

In this work, we assume that the world physics is known to both agents, but the individual capabilities are not mutually-known, i.e., agents know their own abilities, but are unaware of the capabilities of other agents. 

\subsection{Modeling Uncertainty over Capabilities}

Since capabilities are unknown, our POMDP is not fully-specified. We overcome this issue by modeling the \emph{uncertainty} over potential agent capabilities. We adopt the BA-POMDP approach, which enables our robot to consider this uncertainty during \emph{planning}, and \emph{learn} the capabilities from interaction experience. 

Given binary outcomes, the agent capabilities   $P(\tilde{a}| a)$ can be modeled as  Bernoulli distributions (or more generally, categorical distributions for $K$ discrete action outcomes) with unknown parameters $\mathbf{p}$. For each capability distribution, we apply a Dirichlet prior with parameters $\boldsymbol\chi \in \mathcal{X}$ and thus, 
\begin{align}
	P(\tilde{a}|a; \boldsymbol{\chi} ) &= \mathbb{E}_{\mathbf{p}\sim \textrm{Dir}(\boldsymbol{\chi}_a)} [P(\tilde{a}|a, \mathbf{p})] \nonumber \\
	& = \frac{\boldsymbol\chi^{\tilde{a}}_{a}}{\sum_{\tilde{a}'}\boldsymbol\chi^{\tilde{a}'}_{a}} 
	\label{eqn:dirichlet}
\end{align}  
where $\boldsymbol\chi^{\tilde{a}}_{a}$ is an ``experience count'' representing the number of times $\tilde{a}$ occurs when action $a$ is taken. This expectation  is computationally efficient and simple to compute.

As with general BA-POMDPs, the parameters $\boldsymbol\chi$ are embedded into the state. We construct a new POMDP with augmented state set $\hat{S} = S \times \mathcal{X}$. The transition probabilities for the original state $s$ follow Eqn. (\ref{eqn:Tprob}) above, and $\boldsymbol{\chi}$ is deterministically updated via,
\begin{align}
	\boldsymbol{\chi}_{t+1} = \boldsymbol{\chi}_t + \Delta_a^{\tilde{a}}
	\label{eqn:chiupdate}
\end{align}
where $\Delta_a^{\tilde{a}}$ is simply a one hot vector with value 1 at the position corresponding to the $\langle a, \tilde{a}\rangle$. As long as the robot can observe the action and action outcome (or some function of these two elements), it can update the capability model. 

\subsection{Calibrating Intent and Capabilities}
Let us now apply the ideas developed above to human-robot collaboration. We denote the human and robot actions as $a^\text{H}$ and $a^\text{R}$, respectively. Given both actions, the world state transitions according to $P(s_{t+1}|s_t, a^\text{H}, a^\text{R})$. Recall that we seek a policy for our robot whose objective is to help the human accomplish her goal. However, the robot is uncertain about both the intent and capabilities of its human partner. Moreover, the human may be uncertain about the robot's capabilities. 

\mypara{Calibrating Intention.} We assume that the POMDP reward function is parameterized by $\theta \in \Theta$ where $\Theta$ is a finite set of possible rewards. In our scenario, the intention of the human (represented by $\theta$)  is unknown to the robot. Similar to other POMDP-based human-robot collaboration models~\cite{zheng2018pomdp, gopalan2015modeling}, a straightforward way to enable the robot to learn $\theta$ is to embed it into the state---we augment the POMDP's state with $\theta$. Although $\theta$ is latent, human behavior is informative about its value; the human is assumed to act consistently to achieve her objectives. In this work, both human actions (and the corresponding action outcomes) are observable by the robot, and the human follows a fixed policy---this last assumption avoids recursive planning and future work may look into alternative approximations (e.g.,~\cite{malik2018efficient}). Regardless, learning the human's intention without understanding the human's capability may lead to incorrect estimates. 

\mypara{Calibrating Human and Robot Capabilities.} Let $\boldsymbol{\psi} \in \Psi$ be a set of Dirichlet parameters representing the human's capability. The robot maintains a belief over the human's capability by incorporating $\boldsymbol{\psi}$ into the state representation. During state transitions, $\boldsymbol{\psi}$ is updated similar to Eqn. (\ref{eqn:chiupdate}). 

We also calibrate the human's belief over the robot's capability, i.e., we model the human's belief over the robot's capability, and compare it against the robot's true capabilities. We assume that the human is Bayes rational and represent the human's belief over the robot's capability as another set of Dirichlet parameters $\boldsymbol\phi \in \Phi$. 
Similar to the human capability parameters $\boldsymbol{\psi}$, the robot capability belief parameters $\boldsymbol\phi$ are also incorporated into the state, with the transition probabilities adjusted in a similar fashion.

Putting all these components together results in an augmented state space $\bar{S} = S \times {\Theta} \times \Psi \times \Phi$. The transition probability is given by,
\begin{align}
& \bar{T}(\bar{s}_t, a^\text{R}_t, \bar{s}_{t+1}) 
 = \sum_{a^\text{H}} P(\bar{s}_{t+1} | \bar{s}_t, a^\text{R}_t, a^\text{H}) P(a^\text{H} |\bar{s}_t, a^\text{R}_t) \nonumber \\
& = \sum_{\tilde{a}^{\text{R}}} \sum_{\tilde{a}^{\text{H}}} \sum_{a^\text{H}} \Bigg[ P(s_{t+1} | s_t, \tilde{a}^{\text{H}}, \tilde{a}^{\text{R}}) P^*(\tilde{a}^{\text{R}} | a^\text{R}_t)P(\tilde{a}^{\text{H}} | a^\text{H}; \psi_t) \nonumber \\
&  \qquad P(a^\text{H} |\bar{s}_t, a^\text{R}_t)\delta_{\theta_{t+1}, \theta_t} \delta_{\psi_{t+1}, \psi_t + \Delta^{\tilde{a}^{\text{H}}}_{a^\text{H}}} \delta_{\phi_{t+1}, \phi_t + \Delta^{\tilde{a}^{\text{R}}}_{a^\text{R}}} \Bigg]
\end{align}
where $\delta_{i,j}$ is the Kronecker's delta function, which is 1 if $i=j$, and 0 otherwise. Note that unlike the capability parameters, the intention parameter $\theta$ does not change; a different objective is chosen at the beginning of episode and remains fixed until the next episode. 

\mypara{Teaching Actions.}  In order to better facilitate pedagogy behaviour, we expand our action set with failure actions $\hat{a}^\text{F}$, and allow both the human and robot to deliberately choose failure actions when they wished to indicate (in)capability. Performing $\hat{a}^\text{F}$ will always lead to the outcome action $a^\text{F}$. Note that in practice, it may be desirable for the robot to avoid actual failures; rather, choosing $\hat{a}^\text{F}$ would cause the robot to perform a behavior that is indicative of its incapability~\cite{kwon2018expressing}. As such,  $\hat{a}^\text{F}$ can be regarded as a communicative action.

\mypara{Task and Calibration Rewards.} Apart from its primary objective of achieving the task goal, we incentivize the robot to calibrate $\boldsymbol\phi$ to its \emph{actual} capabilities. This is achieved via an additional \emph{calibration reward} at each time-step. This calibration reward is computed based on the similarity between $P(\tilde{a}^{R} | a^\text{R}; \phi)$ and the true capabilities $P^*(\tilde{a}^{R} | a^\text{R})$. We tested several similarity measures and found a simple area overlap (AO) to work well; given two categorical distribution parameters $\mathbf{p}$ and $\mathbf{q}$, we sum  the minimum of the corresponding components of the two distributions, i.e., $\sum_i \min(p_i, q_i)$. We hypothesized the calibration reward would lead to more teaching behaviors, i.e., the robot would attempt to teach the human about its (in)capabilities. This could enable better calibration, which would in-turn lead to better team performance over \emph{multiple} episodes or interaction rounds.
\subsection{Solver Design: TICC-MCP}
In order to solve the TICC-POMDP, we developed an online planning approach we call TICC-MCP (Algorithm 1 and Fig. 1 in the Online Appendix). Online planning methods have been shown to scale well to large state spaces, and the incorporated models allow us to exploit structure and prior knowledge. Our algorithm is similar to the BA-POMCP, with the following two differences:

\mypara{Keeping Additional Statistics.}
In order to simulate $a^\text{H}$, we need to compute $P(a^\text{H}_t | \bar{s}_t, a^\text{R}_t)$ in the search tree. Given the history $h$ at the parent node and the history $ha^\text{R} a^\text{H}$ at the current node, we keep an additional vector of statistics $V_\theta (ha^\text{R} a^\text{H})$ (at each node), which represents the Q-values for the human under different $\theta$. Additionally, we keep $N_\theta (ha^\text{R} a^\text{H})$ and $N_\theta(ha^\text{R})$, which represent the number of times $ha^\text{R} a^\text{H}$ and $ha^\text{R}$ is visited for each $\theta$. These are used to compute the Upper Confidence Bound (UCB1) \cite{lai1985ucb1} values for sampling the human actions. The backup rule for $V_\theta (ha^\text{R} a^\text{H})$, $N_\theta (ha^\text{R} a^\text{H})$ and $N_\theta(ha^\text{R})$ is akin to that for $V(ha^\text{R})$ and $N(ha^\text{R})$.

\mypara{Sampling Human and Robot Actions.} At each node, probability of action $a^\text{H}$ is computed using UCB1~\cite{auer2002finite}:
\begin{equation}
\begin{split}
P(a^\text{H}) & = P\left(a^\text{H} \Big\vert V_\theta (ha^\text{R} a^\text{H}) + c \sqrt{\frac{\log{N_\theta(ha^\text{R})}}{N_\theta (ha^\text{R} a^\text{H})}}\right)
\end{split}
\end{equation}
and the action outcome $\tilde{a}^\text{H}$ is sampled from the human capability model,
$\tilde{a}^\text{H} \sim \sum_{a^\text{H}} P(\tilde{a}^\text{H} | a^\text{H}; \psi) P(a^\text{H})$.
Robot actions are similarly sampled with the appropriate histories, and action outcomes from $P^*(\tilde{a}^\text{R} | a^\text{R})$.
\subsection{Comparison to Related Work}

The TICC-POMDP (and associated solver) is related to a body of literature applying POMDPs towards human robot collaboration (HRC). For example, recent work~\cite{chen2020trust,chen2018planning} has developed trust-based POMDPs that enable robots to consider the human's underlying trust in the robot during planning. Other works have contributed technical innovations for dealing with uncertainty and the complexity of HRC POMDPs, e.g., a Bayesian method to learn the state space and estimate the observation/transition functions~\cite{zheng2018pomdp}, and  hierarchical structure to reduce planning complexity~\cite{ferrari2017cooperative}. Similar to our solution technique, \cite{gopalan2015modeling} modified the vanilla POMCP for large observation spaces associated with HRC POMDPs. Note that the above works do not consider \emph{both} human and robot capabilities and their relationship to intent inference, which is a unique aspect of this paper. 

The TICC-POMDP is similar to Cooperative Inverse Reinforcement Learning (CIRL)~\cite{hadfield2016cooperative} in that both TICC-POMDP and CIRL assume a collaborative setting where the human and robot share the same reward function whose parameters is only known to the human. However, behaviour in CIRL is defined by a joint robot-human policy, while behaviour in TICC-POMDP is defined solely by robot policy with an internal human model. TICC-POMDP relaxes the assumption on human rationality (also considered in ~\cite{malik2018efficient}) and incorporates explicit capability models. 

Our work is also related to methods that model robot (in)capability. For example, recent work has proposed conceptual HRC models based on Theory of Mind~\cite{hellstrom2018understandable},  planning with robot capability models~\cite{sreedharanexpectation}, and a game-theoretic approach for teaching humans about robot capabilities~\cite{nikolaidis2017game}. Very little work has looked into incorporating human capability models into HRC; a possible  exception~\cite{zhang2015capability} modelled human capabilities with Bayesian Networks for planning.
\begin{figure*}
    \centering
    \resizebox{0.8\textwidth}{!}{
    \input{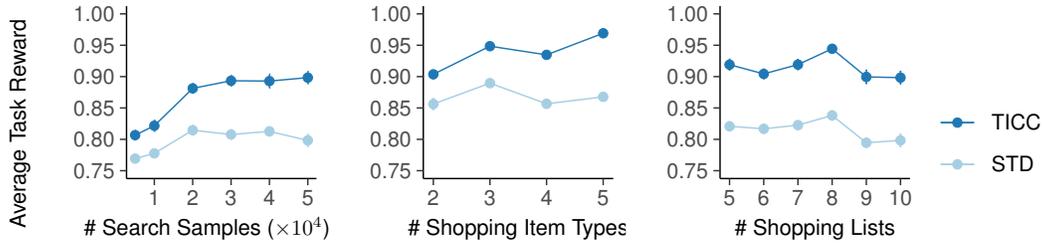}
    }
    \vspace{-2mm}
    \caption{The average task rewards obtained by TICC-MCP and standard POMCP during evaluation stage with varying number of search samples (left), shopping item types (middle) and shopping lists (right). Error bars indicate one standard error.}
    \label{fig:results}
    \vspace{-3mm}
\end{figure*}

\section{Simulation Experiments}

We evaluated our model using both simulated and real-world experiments. Our main hypothesis was that calibrating a human capability model resulted in higher task rewards over the long term, i.e., over multiple episodes or interaction rounds. We compared the TICC-MCP (TICC) against a standard POMCP algorithm (STD). The standard POMCP performed intent inference but did not explicitly model capability nor had a reward calibration component. 

This section describes our simulation experiments where we focused on evaluating the algorithmic aspects of our approach. We were interested in whether our method was able to infer the intent and capabilities of a simulated human, under various parameter settings.

\mypara{Domain.}
Our task was in the shopping domain where the human and robot collaborate to collect a bag of items within a limited horizon. There are $n$  item types and $m$ possible shopping lists, and the world state is a $n$-tuple representing the quantity of each item already collected. Although seemingly abstract and small, the problem described above is challenging and similar to domains explored in recent work~\cite{malik2018efficient}. The largest of our experiments involved more than $10^{10}$ world states. 

The human and robot would collaborate over multiple episodes (rounds). At the beginning of each round, the human observes the intended shopping list, which corresponds to a specific reward parameter $\theta$. Only the human knows the shopping list at the outset, while the robot starts with a uniform belief over each list. Both agents are rewarded based on how closely their item bag matches the goal list. At each time-step, the robot and human each chooses to pick-up an item, or does nothing. Each item type can be chosen multiple times throughout each episode (there is an unlimited number of items).  Neither the human nor the robot are perfectly capable and may fail to pick up certain items.  

\mypara{Manipulated Variables.} To evaluate  our method, we varied the number of search samples, shopping item types (i.e., the size of action and observation spaces), and shopping lists (the size of the reward space).

\mypara{Dependent Measures.}
We evaluated the performance of the two algorithms using the the task reward received and the correctness of the belief over the human's intent $\theta$ and human's capability $\psi$. The correctness of the intent was measured by average likelihood of the true $\theta$ given the model. Capability correctness was measured as the area of overlap between the normalized belief distribution and the true capability distribution. The measures were computed over 50 independent runs.

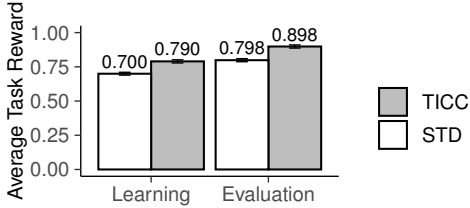
\begin{figure} 
    \centering 
    \resizebox{0.8\linewidth}{!}{
    \begin{tikzpicture}[x=1pt,y=1pt,font=\sffamily]
\definecolor{fillColor}{RGB}{255,255,255}
\path[use as bounding box,fill=fillColor,fill opacity=0.00] (0,0) rectangle (216.81,119.25);
\begin{scope}
\path[clip] (  0.00,  0.00) rectangle (216.81,119.25);
\definecolor{drawColor}{RGB}{255,255,255}
\definecolor{fillColor}{RGB}{255,255,255}

\path[draw=drawColor,line width= 0.6pt,line join=round,line cap=round,fill=fillColor] (  0.00,  0.00) rectangle (216.81,119.25);
\end{scope}
\begin{scope}
\path[clip] ( 38.56, 30.69) rectangle (151.39, 96.59);
\definecolor{fillColor}{RGB}{255,255,255}

\path[fill=fillColor] ( 38.56, 30.69) rectangle (151.39, 96.59);
\definecolor{drawColor}{RGB}{0,0,0}
\definecolor{fillColor}{RGB}{190,190,190}

\path[draw=drawColor,line width= 0.9pt,line cap=rect,fill=fillColor] ( 69.33, 33.68) rectangle ( 92.41, 80.99);
\definecolor{fillColor}{RGB}{255,255,255}

\path[draw=drawColor,line width= 0.9pt,line cap=rect,fill=fillColor] ( 46.25, 33.68) rectangle ( 69.33, 75.59);
\definecolor{fillColor}{RGB}{190,190,190}

\path[draw=drawColor,line width= 0.9pt,line cap=rect,fill=fillColor] (120.62, 33.68) rectangle (143.70, 87.50);
\definecolor{fillColor}{RGB}{255,255,255}

\path[draw=drawColor,line width= 0.9pt,line cap=rect,fill=fillColor] ( 97.54, 33.68) rectangle (120.62, 81.51);

\path[draw=drawColor,line width= 0.6pt,line join=round] ( 78.30, 81.69) --
	( 83.43, 81.69);

\path[draw=drawColor,line width= 0.6pt,line join=round] ( 80.87, 81.69) --
	( 80.87, 80.29);

\path[draw=drawColor,line width= 0.6pt,line join=round] ( 78.30, 80.29) --
	( 83.43, 80.29);

\path[draw=drawColor,line width= 0.6pt,line join=round] ( 55.22, 76.23) --
	( 60.35, 76.23);

\path[draw=drawColor,line width= 0.6pt,line join=round] ( 57.79, 76.23) --
	( 57.79, 74.96);

\path[draw=drawColor,line width= 0.6pt,line join=round] ( 55.22, 74.96) --
	( 60.35, 74.96);

\path[draw=drawColor,line width= 0.6pt,line join=round] (129.60, 88.21) --
	(134.72, 88.21);

\path[draw=drawColor,line width= 0.6pt,line join=round] (132.16, 88.21) --
	(132.16, 86.80);

\path[draw=drawColor,line width= 0.6pt,line join=round] (129.60, 86.80) --
	(134.72, 86.80);

\path[draw=drawColor,line width= 0.6pt,line join=round] (106.51, 82.19) --
	(111.64, 82.19);

\path[draw=drawColor,line width= 0.6pt,line join=round] (109.08, 82.19) --
	(109.08, 80.83);

\path[draw=drawColor,line width= 0.6pt,line join=round] (106.51, 80.83) --
	(111.64, 80.83);

\node[text=drawColor,anchor=base,inner sep=0pt, outer sep=0pt, scale=  0.8] at ( 80.87, 83.59) {0.790};

\node[text=drawColor,anchor=base,inner sep=0pt, outer sep=0pt, scale=  0.8] at ( 57.79, 78.20) {0.700};

\node[text=drawColor,anchor=base,inner sep=0pt, outer sep=0pt, scale=  0.8] at (132.16, 90.11) {0.898};

\node[text=drawColor,anchor=base,inner sep=0pt, outer sep=0pt, scale=  0.8] at (109.08, 84.11) {0.798};
\end{scope}
\begin{scope}
\path[clip] (  0.00,  0.00) rectangle (216.81,119.25);
\definecolor{drawColor}{RGB}{0,0,0}

\path[draw=drawColor,line width= 0.6pt,line join=round] ( 38.56, 30.69) --
	( 38.56, 96.59);
\end{scope}
\begin{scope}
\path[clip] (  0.00,  0.00) rectangle (216.81,119.25);
\definecolor{drawColor}{gray}{0.30}

\node[text=drawColor,anchor=base east,inner sep=0pt, outer sep=0pt, scale=  0.88] at ( 33.61, 30.65) {0.00};

\node[text=drawColor,anchor=base east,inner sep=0pt, outer sep=0pt, scale=  0.88] at ( 33.61, 45.63) {0.25};

\node[text=drawColor,anchor=base east,inner sep=0pt, outer sep=0pt, scale=  0.88] at ( 33.61, 60.61) {0.50};

\node[text=drawColor,anchor=base east,inner sep=0pt, outer sep=0pt, scale=  0.88] at ( 33.61, 75.58) {0.75};

\node[text=drawColor,anchor=base east,inner sep=0pt, outer sep=0pt, scale=  0.88] at ( 33.61, 90.56) {1.00};
\end{scope}
\begin{scope}
\path[clip] (  0.00,  0.00) rectangle (216.81,119.25);
\definecolor{drawColor}{gray}{0.20}

\path[draw=drawColor,line width= 0.6pt,line join=round] ( 35.81, 33.68) --
	( 38.56, 33.68);

\path[draw=drawColor,line width= 0.6pt,line join=round] ( 35.81, 48.66) --
	( 38.56, 48.66);

\path[draw=drawColor,line width= 0.6pt,line join=round] ( 35.81, 63.64) --
	( 38.56, 63.64);

\path[draw=drawColor,line width= 0.6pt,line join=round] ( 35.81, 78.61) --
	( 38.56, 78.61);

\path[draw=drawColor,line width= 0.6pt,line join=round] ( 35.81, 93.59) --
	( 38.56, 93.59);
\end{scope}
\begin{scope}
\path[clip] (  0.00,  0.00) rectangle (216.81,119.25);
\definecolor{drawColor}{RGB}{0,0,0}

\path[draw=drawColor,line width= 0.6pt,line join=round] ( 38.56, 30.69) --
	(151.39, 30.69);
\end{scope}
\begin{scope}
\path[clip] (  0.00,  0.00) rectangle (216.81,119.25);
\definecolor{drawColor}{gray}{0.20}

\path[draw=drawColor,line width= 0.6pt,line join=round] ( 69.33, 27.94) --
	( 69.33, 30.69);

\path[draw=drawColor,line width= 0.6pt,line join=round] (120.62, 27.94) --
	(120.62, 30.69);
\end{scope}
\begin{scope}
\path[clip] (  0.00,  0.00) rectangle (216.81,119.25);
\definecolor{drawColor}{gray}{0.30}

\node[text=drawColor,anchor=base,inner sep=0pt, outer sep=0pt, scale=  0.88] at ( 69.33, 19.68) {Learning};

\node[text=drawColor,anchor=base,inner sep=0pt, outer sep=0pt, scale=  0.88] at (120.62, 19.68) {Evaluation};
\end{scope}
\begin{scope}
\path[clip] (  0.00,  0.00) rectangle (216.81,119.25);
\definecolor{drawColor}{RGB}{0,0,0}

\node[text=drawColor,rotate= 90.00,anchor=base,inner sep=0pt, outer sep=0pt, scale=  0.9] at ( 13.08, 63.64) {Average Task Reward};
\end{scope}
\begin{scope}
\path[clip] (  0.00,  0.00) rectangle (216.81,119.25);
\definecolor{fillColor}{RGB}{255,255,255}

\path[fill=fillColor] (162.39, 36.08) rectangle (211.31, 91.20);
\end{scope}
\begin{scope}
\path[clip] (  0.00,  0.00) rectangle (216.81,119.25);
\definecolor{drawColor}{RGB}{0,0,0}
\definecolor{fillColor}{RGB}{190,190,190}

\path[draw=drawColor,line width= 0.9pt,line cap=rect,fill=fillColor] (169.03, 57.17) rectangle (181.21, 69.35);
\end{scope}
\begin{scope}
\path[clip] (  0.00,  0.00) rectangle (216.81,119.25);
\definecolor{drawColor}{RGB}{0,0,0}
\definecolor{fillColor}{RGB}{255,255,255}

\path[draw=drawColor,line width= 0.9pt,line cap=rect,fill=fillColor] (169.03, 42.71) rectangle (181.21, 54.89);
\end{scope}
\begin{scope}
\path[clip] (  0.00,  0.00) rectangle (216.81,119.25);
\definecolor{drawColor}{RGB}{0,0,0}

\node[text=drawColor,anchor=base west,inner sep=0pt, outer sep=0pt, scale=  0.88] at (187.85, 60.23) {TICC};
\end{scope}
\begin{scope}
\path[clip] (  0.00,  0.00) rectangle (216.81,119.25);
\definecolor{drawColor}{RGB}{0,0,0}

\node[text=drawColor,anchor=base west,inner sep=0pt, outer sep=0pt, scale=  0.88] at (187.85, 45.77) {STD};
\end{scope}
\end{tikzpicture}
    }
    \vspace{-8mm}
    \caption{The average task rewards over simulation rounds. Error bars indicate one standard error.}
    \label{fig:taskrewards}
    \vspace{-2mm}
\end{figure}

\begin{figure} 
    \centering 
    \resizebox{0.8\linewidth}{!}{
    \input{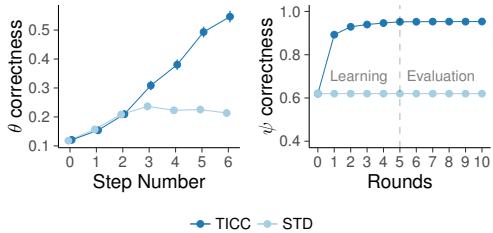}
    }
    \caption{$\theta$ belief correctness over number of steps taken in a single episode (left). $\psi$ belief correctness of TICC-MCP and standard POMCP over simulation rounds (right). Error bars indicate one standard error.}
    \label{fig:calib}
    \vspace{-3mm}
\end{figure}

\mypara{Method.} The robot was initialized to believe that the human is fully capable. Each experiment consisted of two stages: a learning stage and an evaluation stage. In the learning stage, we first ran 5 rounds of simulation for the robot to learn about the human's capability. In the evaluation stage, we ran another 5 rounds, from which the average task reward was calculated. Note that the human capability parameters $\boldsymbol\psi$ carry over the rounds, but $\theta$ is initialized randomly at the beginning of each round. 

\mypara{Results.} In brief, the simulation results indicate that TICC-MCP agent was better able to infer the human's intent and capability, which led to improved task rewards over the long-term. We see in Fig.~\ref{fig:results} that our approach consistently achieved higher average task rewards across the variable changes. Fig. \ref{fig:taskrewards} and Fig. \ref{fig:calib} show an illustrative result with 10 shopping lists, 5 items, and 50k search samples. TICC was able to quickly learn the human's capabilities. It was better able to estimate the correct $\theta$ leading to higher performance compared to STD, particularly in the evaluation stage.
\section{Human Subject Experiments}
This section describes our human subject experiments where we sought to assess the TICC-MCP. Our primary hypotheses were that:
\begin{itemize}
\item[\textbf{H1}] Calibrated intent and capability models improves performance (i.e. task rewards) over a sequence of episodes.
\item[\textbf{H2}] Mutual calibration induces higher trust in the assistive robot.
\end{itemize}

\mypara{Experimental Setup.}
Participants interacted with a real Fetch robot in the collaborative item-shopping task in a table-top setting (see Fig. \ref{fig:human_exp_setup}). Four types of items were available for purchase: yellow cups, sweets, corrosive cleaning agents and green cups. Half of the yellow cups were ``faulty'' (marked with a cross) while the other half are ``good'' (marked with a tick); note that the markings were only visible under UV light. This simulates a grocery shopping scenario where it may be difficult to distinguish between good and faulty products with the naked eye (e.g., good or bad fruits may be hard to distinguish).
In our experiment, the human is unable to pick up the corrosive cleaning agent, and the robot is unable to pick up the small sweets. The human has a 50\% chance of picking a ``good'' yellow cup since half of them are faulty. In contrast, the robot has an 80\% probability of picking a ``good'' yellow cup.

The horizon was fixed at 6 steps for both agents. At each step, the robot will make the first move followed by the human. Each of the shopping lists had either 8 or 9 items. As such, the human participant is unable to single-handedly fulfil the shopping list due to her imperfect capability and the limited horizon. Each experiment lasted $\approx 60$ minutes.

\vspace{-1mm}
\mypara{Procedure.}
Participants entered the lab and were briefed about the task. They were also told that they have two types of privileged information: 1) their own capability and 2) the actual shopping list for that round. They then engaged in a practice round where a human demonstrator took the place of the Fetch robot. This was done to avoid carryover effects from the practice to the main experiment.

\begin{figure}
\centering
\subfloat[]{
\includegraphics[width=0.40\linewidth]{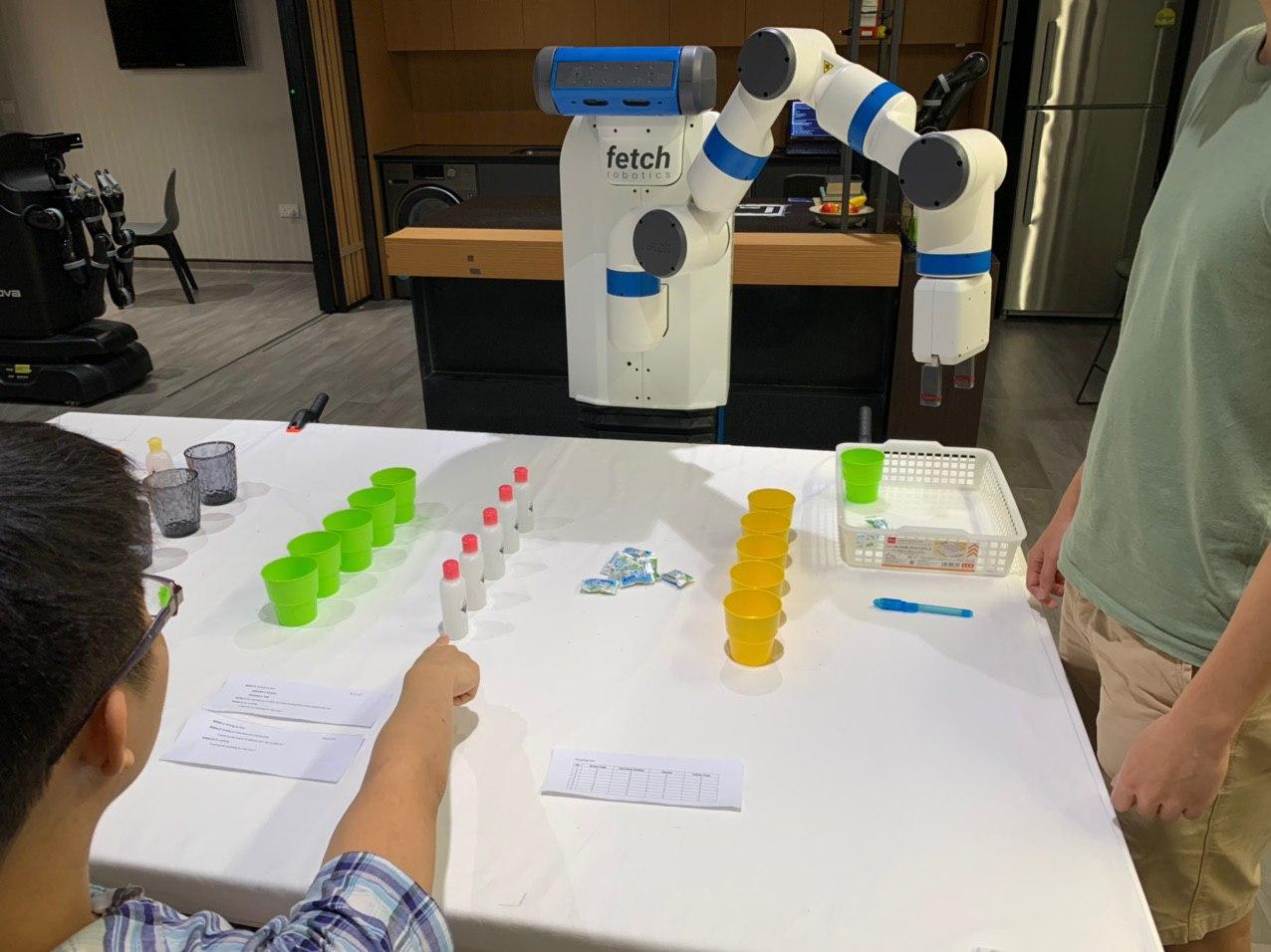}
\label{fig:human_fails_bottle}
}
\subfloat[]{
\includegraphics[width=0.40\linewidth]{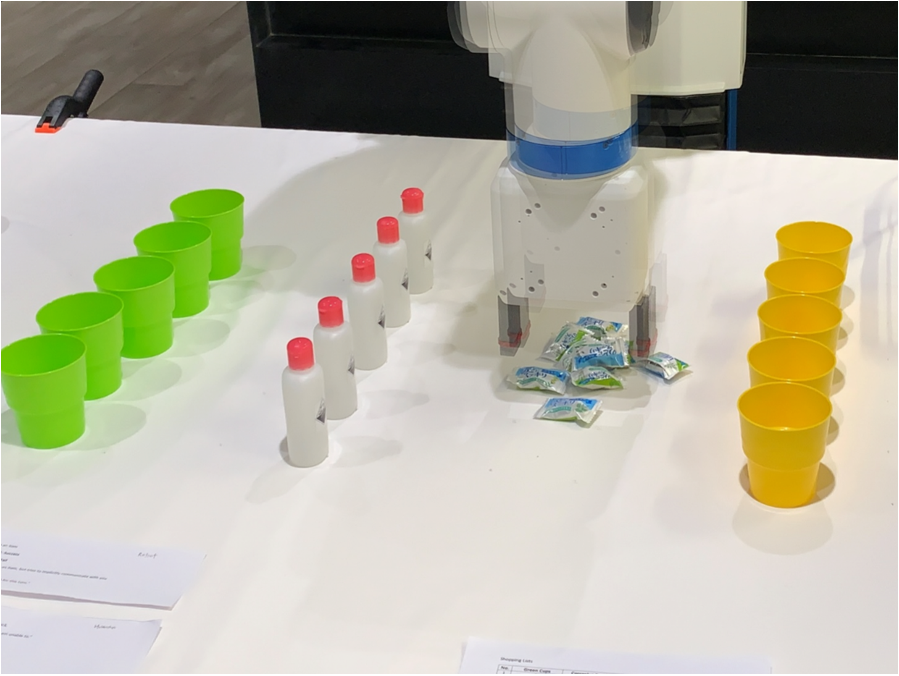}
\label{fig:robot_fails_sweets}
}
\caption{Examples of communicative actions. (a) The human indicates to the robot that he wants to pick the corrosive cleaning agent, but is unable to; (b) The robot executes a communicative action with the participant about its capability by hovering above the sweets.}
\label{fig:robot_action_space}
\vspace{-5mm}
\end{figure}

Thereafter, they performed 5 consecutive rounds of the task with the real robot. A different shopping list, selected by a seeded random number generator, was used for each round. The order of the shopping list was fixed for all participants to reduce experimental variability and to allow for a better comparison between the two algorithms. Participants received $\$1$ or $\$0.50$ bonus if they completed the shopping list or if the number of items exceeded what was required respectively. On every turn, both the participant and robot can choose to either pick an item or do nothing. Additionally, participants can choose to indicate to the robot that they want to pick an item, but is unable to (Fig. \ref{fig:human_fails_bottle}). Likewise, the robot may choose actions to communicate with the participant (Fig. \ref{fig:robot_fails_sweets}).

{\tiny
\begin{table}[]
\caption{\label{tab:subjective_measures} Subjective measures in the human experiment.}
\begin{tabular}{@{}ll@{}}
\toprule
& \textbf{Subjective Measures} \\                       \midrule
\textbf{Before Interaction} & 
\begin{tabular}[c]{@{}l@{}}
\textbf{General Perception of Robot} \\
- Negative Attitudes toward Situations \\and Interactions  with Robots~\cite{syrdal2009negative}  \\
- Schaefer's Trust Perception Scale~\cite{schaefer2016measuring} 
\end{tabular}  \\                                     
\midrule
\textbf{Before $1^{st}$ Round}   & 
\begin{tabular}[c]{@{}l@{}}
\textbf{Perceived Relative Capability} \\
\textit{(Robot is better/worse/both are equally capable)} \\
- Do you think the robot is better than you at \\ 
picking the [insert item name]?
\end{tabular} \\
\midrule
\textbf{After each Round}   & 
\begin{tabular}[c]{@{}l@{}}
\textbf{Perceived Relative Capability} \\
\textit{(Robot is better/worse/both are equally capable)} \\
- Do you think the robot is better than you at \\ 
picking the [insert item name]?\\ 
\\
\textbf{Human-Robot Trust} \\ 
\textit{(5-point Likert Scale)} \\ 
- I trust the robot to collaborate with \\
me on this task \\ 
- The robot trusts me to collaborate with \\
it on this task\\
\\
\textbf{Human's First-Order Belief of Robot's Beliefs} \\
\textit{(5-point Likert Scale)} \\ 
- The robot understands what I am capable of\\ 
- The robot is able to infer my intentions
\end{tabular} \\ 
\midrule
\textbf{After Interaction}  & 
\begin{tabular}[c]{@{}l@{}}
\textbf{General Perception of Robot} \\
- Negative Attitudes toward Situations \\ and Interactions  with Robots \\ 
- Schaefer's Trust Perception Scale
\end{tabular} \\ 
\bottomrule
\end{tabular}
\end{table}
}
\vspace{-1mm}

\mypara{Participant Assignment.}
A total of 28 participants (mean age = 22.8, 16 females) were recruited from the university community. The experiment was designed to be between-subjects with 14 participants in each condition. They were randomly assigned to play with either the TICC-MCP or standard POMCP robot.

\mypara{Dependent Measures.}
We used both objective and subjective measures to evaluate the performance of the algorithms. There were three objective performance measures: 
\begin{itemize}
    \item The average task reward for the final three rounds
    \item The calibration score of belief of robot's capability $\phi$
    \item The calibration score of belief of human's capability $\psi$
\end{itemize}
We also collected the subjective measures shown in Table \ref{tab:subjective_measures} throughout the course of the experiment.

\begin{figure}
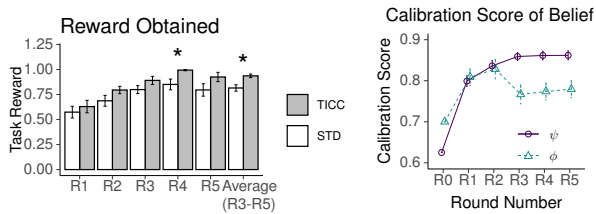
 
    \centering
    \resizebox{0.55\linewidth}{!}{
    \subfloat{\input{images/reward_plot_3}}
    }
    \resizebox{0.35\linewidth}{!}{
    \subfloat{\input{images/chi}}
    }
    \vspace{-3mm}
    \caption{Task rewards over rounds (left). Convergence of belief over both $\psi$ and $\phi$ across simulation rounds for TICC-MCP agent (right) . Error bars indicate one standard error. Plots marked with * indicate statistical significance.}
    \label{fig:zero_order}
\end{figure}

\begin{figure}
    \centering
    \resizebox{0.85\linewidth}{!}{
	\input{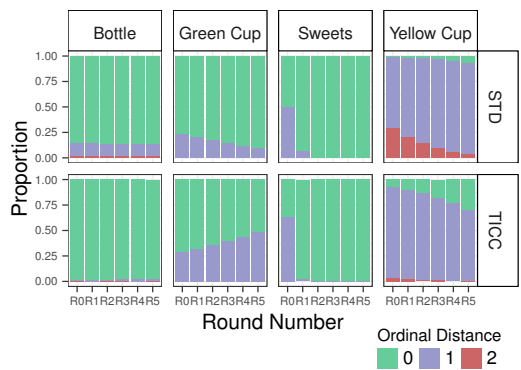}
	}
	\vspace{-13mm}
    \caption{Predicted probabilities (based on fixed-effects estimates) of the ordinal distance at each time point. An ordinal distance of 0 indicates that the participant's beliefs of the robot's capability agrees with reality.}
	\label{fig:ord_plot}
\end{figure}

\mypara{Results.} The data from 4 participants were removed due to a consistency check question failure\footnote{They responded that the robot was worse at picking-up the bottle at the last round, despite the fact that they cannot pick up the bottle.}. All analyses were carried out with available-case analysis.
As seen in Fig. \ref{fig:zero_order}, the mean calibration scores for both $\psi$ and $\phi$ show an increasing trend. The increase in $\psi$ calibration scores suggests that the TICC-MCP robot can effectively learn about the participants' capabilities. The increase in $\phi$ calibration scores suggests that the TICC-MCP robot influenced the participants' belief of its capability. More details can be seen in Fig. \ref{fig:ord_plot} which indicates the participants beliefs over the robot's capabilities gradually matched reality. 

In brief, we found that the TICC-MCP algorithm led to significantly higher average task rewards in the final three rounds (Fig. \ref{fig:zero_order}, $t(22) = 3.474, p = 0.00215$). Separate two-sample t-tests at each of the final three rounds suggest that participants who worked with the TICC-MCP robot achieved higher rewards (i.e. the t-values are all positive) (Round 3: $t(22) = 1.599, p = 0.124$; Round 4: $t(22) = 2.881, p = 0.0087$; Round 5: $t(22) = 1.669, p = 0.103$), although only the difference in Round 4 achieved statistical significance when $\alpha = 0.05$. The difference between the averaged rewards and the reward at Round 4 is statistically significant after adjusting for multiple comparisons using Bonferroni correction (adjusted-$\alpha$ $= 0.0125$). These results support hypothesis \textbf{H1}.

\begin{figure}
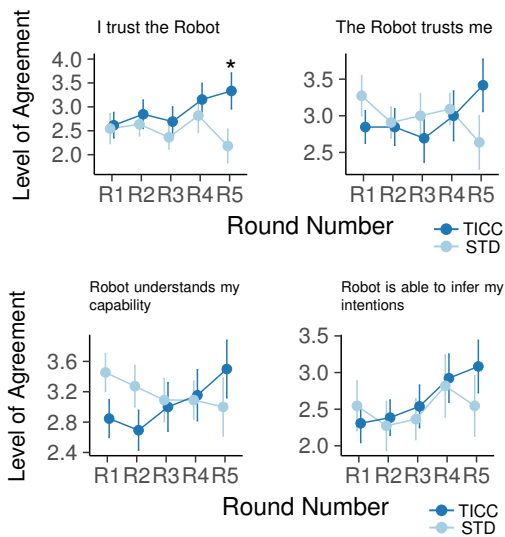

\centering
    \vspace{-8mm}
    \resizebox{0.85\linewidth}{!}{
	\input{images/misc_trust_2.tex}}
	
	\vspace{-10mm}
	\resizebox{0.85\linewidth}{!}{
	\input{images/misc_first_order.tex}}
	\vspace{-5mm}
    \caption{Subjective measures of Human-Robot Trust and the participants' perception of the robot's ability to understand their capabilities and intentions. Error bars indicate one standard error. Plots marked with * indicate statistical significance.}
	\label{fig:misc_trust}
	\vspace{-6mm}
\end{figure}

We carried out planned two-sample comparisons for each of the four survey questions at the end of the final round.
Participants in the TICC-MCP condition reported having higher levels of trust in the robot ($t(21) = 2.224, p = 0.0372$) (see Fig. \ref{fig:misc_trust} (top left)). This provides evidence for hypothesis \textbf{H2}. The differences for other three questions did not achieve statistical significance when $\alpha = 0.05$. Nevertheless, Fig. \ref{fig:misc_trust} suggests that participants in different conditions perceived the robot differently as they interacted with it.

\section{Conclusion}
This paper contributes the TICC-POMDP and associated TICC-MCP online solver, which our experiments show can enable better human-robot cooperation over multiple interactions. Calibrating mutual beliefs of intention and capability resulted in both higher team performance and higher human trust. As future work, we intend to explore applying TICC to longer-term real-world scenarios and with other salient factors, e.g., across multiple tasks~\cite{soh2018transfer,soh2020multi} or where attention is crucial~\cite{bodala2020modeling}.

\section*{Acknowledgements}
This work was supported by the SERC, A*STAR, Singapore, through the National Robotics Program under Grant No. W1925d0054. 

\addtolength{\textheight}{0cm}

\bibliographystyle{IEEEtran}
\bibliography{references}

\section*{APPENDIX}

\subsection{Additional Details for TICC-MCP}
\begin{figure*}
\centering
\includegraphics[width=0.85\textwidth]{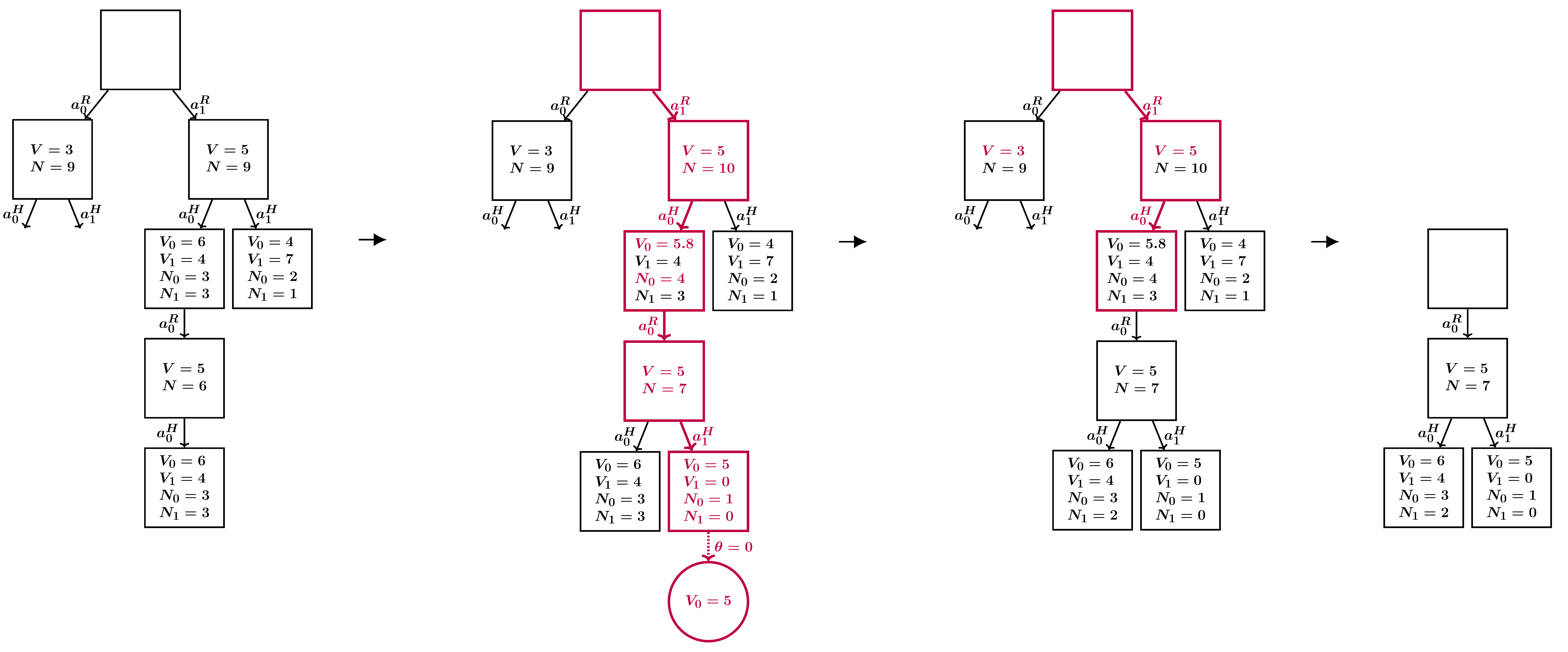}
\caption{An example illustration of TICC-MCP in a simple environment with 2 robot actions $a^\text{R} \in \{a^\text{R}_0, a^\text{R}_1\}$, 2 human actions $a^\text{H} \in \{a^\text{H}_0, a^\text{H}_1\}$, 2 reward parameters $\theta \in \{0, 1\}$ , no intermediate rewards and a discount factor of 1. For brevity, the belief particles at each node are omitted. The robot constructs a search tree from multiple search iterations and stores the values $V$ and number of visits $N$ in every node/history (left). When visiting a new leaf node, the robot performs expansion and simulation using rollout policy, followed by backpropagation (second left). The robot compares the values in the search tree and choose action $a^\text{R}_1$; the human then chooses action $a^\text{H}_0$ (second right). The robot prunes the tree and begins a new search from the updated history $ha^\text{R}_1a^\text{H}_0$ (right).}
\label{fig:HA-POMCP}
\end{figure*}

The pseudocode and a simple running example for the TICC-MCP is shown in Algorithm \ref{HA-POMCP} and Fig. \ref{fig:HA-POMCP} respectively.

\begin{algorithm}
\scriptsize
\linespread{1}\selectfont
\caption{HA-POMCP}\label{HA-POMCP}
\begin{algorithmic}[1]
\Procedure{SEARCH}{$h$}
\Repeat
\If {$h=empty$}
	\State $\bar{s}\sim \cal{I}$
\Else
	\State $\bar{s}\sim B(h)$
	\Call{SIMULATE}{$\bar{s}, h, 0$}
\EndIf
\Until{\Call{TIMEOUT}{}}
\State \Return $\arg\max_{a^\text{R}} V(ha^\text{R})$
\EndProcedure
\\
\Procedure{ROLLOUT}{$\bar{s},h,depth$}
\If {$\gamma ^{depth} < \epsilon$}
	\State \Return $0$
\EndIf
\State $a^\text{R}, a^\text{H} \sim Uniform(a^\text{R}), Uniform(a^\text{H})$
\State $\tilde{a}^\text{R} \sim P^*(\tilde{a}^\text{R} | a^\text{R})$
\State $\tilde{a}^\text{H} \sim P(\tilde{a}^\text{H} | a^\text{H}; \psi)$
\State $\bar{s}' \leftarrow T(\bar{s}, a^\text{R}, a^\text{H}, \tilde{a}^\text{R}, \tilde{a}^\text{H})$
\State \Return $r(\bar{s})+\gamma$ \Call{ROLLOUT}{$\bar{s}',ha^\text{R}a^\text{H},depth+1$}
\EndProcedure
\\
\Procedure{SIMULATE}{$\bar{s},h,depth$}
\If {$\gamma ^{depth} < \epsilon$}
	\State \Return $0$
\EndIf
\If {$h \notin T$}
	\State \Return \Call{ROLLOUT}{$\bar{s},h,depth$}
\EndIf
\State $a^\text{R} \leftarrow \arg\max_{a^\text{R}} V (ha^\text{R}) + c \sqrt{\frac{\log{N(h)}}{N (ha^\text{R})}}$
\State $a^\text{H} \sim  \pi_H(a^\text{H}| V_\theta (ha^\text{R} a^\text{H}) + c\sqrt{\frac{\log{N_\theta(ha^\text{R})}}{N_\theta (ha^\text{R} a^\text{H})}})$
\State $\tilde{a}^\text{R} \sim P^*(\tilde{a}^\text{R} | a^\text{R})$
\State $\tilde{a}^\text{H} \sim P(\tilde{a}^\text{H} | a^\text{H}; \psi)$
\State $\bar{s}' \leftarrow T(\bar{s}, a^\text{R}, a^\text{H}, \tilde{a}^\text{R}, \tilde{a}^\text{H})$
\State $R \leftarrow r(\bar{s}) + \gamma$ \Call{SIMULATE}{$\bar{s}',ha^\text{R}a^\text{H},depth+1$}
\State $B(h) \leftarrow B(h) \cup \{\bar{s}\}$
\State $N(h) \leftarrow N(h)+1$
\State $N_\theta(ha^\text{R}) \leftarrow N_\theta(ha^\text{R}) + 1$
\State $N_\theta(ha^\text{R}a^\text{H}) \leftarrow N_\theta(ha^\text{R}a^\text{H}) + 1$
\State $V(ha^\text{R}) \leftarrow V(ha^\text{R}) + \frac{R-V(ha^\text{R})}{N(ha^\text{R})}$
\State $V_\theta(ha^\text{R}a^\text{H}) \leftarrow V_\theta(ha^\text{R}a^\text{H}) + \frac{R-V_\theta(ha^\text{R}a^\text{H})}{N_\theta(ha^\text{R}a^\text{H})}$
\State \Return $R$
\EndProcedure
\end{algorithmic}
\end{algorithm}

\subsection{Additional Details for Simulation Experiments}
In all simulation experiments, there were 12 or 13 shopping items to purchase in each shopping list. The shopping lists used for the experiments were created randomly at the start and fixed thereafter. The TICC-MCP robot initially assumes that 1) the human has perfect capability 2) the human thinks that the robot has perfect capability.

\mypara{Setup 1: Varying Number of Search Samples}
In this setup, the number of shopping lists was fixed at 10 and number of shopping item types was fixed at 5. The number of search samples was varied from 5000 to 50000. The shopping lists and the actual capabilities setup are shown in Table \ref{tab:setup1} respectively.

\begin{table}[ht]
\small
\caption{Shopping lists (top), human capability in success rate (middle) and robot capability in success rate (bottom) used for simulation experiment Setup 1.}
\begin{center}
    {Shopping Lists \\}
    \begin{tabular}{|c|c|c|c|c|} 
    \hline
    \textbf{Item 1} & \textbf{Item 2} & \textbf{Item 3}  & \textbf{Item 4} & \textbf{Item 5}\\ 
    \hline
    4 & 3 & 0 & 2 & 3 \\ 
    \hline
    1 & 4 & 0 & 7 & 1 \\
    \hline
    2 & 3 & 2 & 3 & 3 \\
    \hline
    5 & 4 & 2 & 0 & 2 \\
    \hline
    0 & 3 & 3 & 4 & 3 \\
    \hline
    3 & 3 & 0 & 3 & 3 \\
    \hline
    6 & 3 & 0 & 1 & 2 \\
    \hline
    2 & 3 & 4 & 1 & 2 \\
    \hline
    1 & 1 & 2 & 4 & 4 \\
    \hline
    0 & 3 & 2 & 5 & 2 \\
    \hline
    \end{tabular}
\end{center}

\begin{center}
    {Human's capability \\}
    \begin{tabular}{|c|c|c|c|c|} 
    \hline
    \textbf{Item 1} & \textbf{Item 2} & \textbf{Item 3}  & \textbf{Item 4} & \textbf{Item 5}\\
    \hline
    0\% & 100\% & 10\% & 0\% & 100\% \\
    \hline
    \end{tabular}
\end{center}

\begin{center}
    {Robot's capability \\}
    \begin{tabular}{|c|c|c|c|c|} 
    \hline
    \textbf{Item 1} & \textbf{Item 2} & \textbf{Item 3}  & \textbf{Item 4} & \textbf{Item 5}\\
    \hline
    100\% & 0\% & 100\% & 100\% & 10\% \\
    \hline
    \end{tabular}
\label{tab:setup1}
\end{center}
\end{table}

\mypara{Setup 2: Varying Number of Shopping Item Types}
In this setup, the number of shopping lists was fixed at 2 and number of search samples was fixed at 50000. The number of shopping item types was varied from 2 to 5. The shopping lists and the actual capabilities setups for 2, 3, 4 and 5 item types are shown in Tables \ref{tab:setup2.2}, \ref{tab:setup2.3}, \ref{tab:setup2.4} and \ref{tab:setup2.5} respectively.

\begin{table}[ht]
\small
\caption{Shopping lists (top), human capability in success rate (middle) and robot capability in success rate (bottom) used for simulation experiment Setup 2 with 2 shopping item types.}
\begin{center}
    {Shopping lists \\}
    \begin{tabular}{|c|c|} 
    \hline
    \textbf{Item 1} & \textbf{Item 2}\\ 
    \hline
    1 & 12 \\ 
    \hline
    2 & 10 \\
    \hline
    \end{tabular}
\end{center}

\begin{center}
    {Human's capability \\}
    \begin{tabular}{|c|c|} 
    \hline
    \textbf{Item 1} & \textbf{Item 2}\\ 
    \hline
    50\% & 100\% \\
    \hline
    \end{tabular}
\end{center}

\begin{center}
    {Robot's capability \\}
    \begin{tabular}{|c|c|} 
    \hline
    \textbf{Item 1} & \textbf{Item 2}\\ 
    \hline
    100\% & 50\%\\
    \hline
    \end{tabular}
\end{center}
\label{tab:setup2.2}
\end{table}

\begin{table}[ht]
\small
\caption{Shopping lists (top), human capability in success rate (middle) and robot capability in success rate (bottom) used for simulation experiment Setup 2 with 3 shopping item types.}
\begin{center}
    {Shopping lists \\}
    \begin{tabular}{|c|c|c|} 
    \hline
    \textbf{Item 1} & \textbf{Item 2} & \textbf{Item 3}\\ 
    \hline
    8 & 5 & 0 \\ 
    \hline
    2 & 5 & 6 \\
    \hline
    \end{tabular}
\end{center}

\begin{center}
    {Human's capability \\}
    \begin{tabular}{|c|c|c|} 
    \hline
    \textbf{Item 1} & \textbf{Item 2} & \textbf{Item 3}\\ 
    \hline
    0\% & 100\% & 10\% \\
    \hline
    \end{tabular}
\end{center}

\begin{center}
    {Robot's capability \\}
    \begin{tabular}{|c|c|c|} 
    \hline
    \textbf{Item 1} & \textbf{Item 2} & \textbf{Item 3}\\ 
    \hline
    100\% & 0\% & 100\% \\
    \hline
    \end{tabular}
\end{center}
\label{tab:setup2.3}
\end{table}

\begin{table}[ht]
\small
\caption{Shopping lists (top), human capability in success rate (middle) and robot capability in success rate (bottom) used for simulation experiment Setup 2 with 4 shopping item types.}
\begin{center}
    {Shopping lists \\}
    \begin{tabular}{|c|c|c|c|} 
    \hline
    \textbf{Item 1} & \textbf{Item 2} & \textbf{Item 3} & \textbf{Item 4}\\ 
    \hline
    4 & 4 & 2 & 3 \\ 
    \hline
    3 & 5 & 0 & 5 \\
    \hline
    \end{tabular}
\end{center}

\begin{center}
    {Human's capability \\}
    \begin{tabular}{|c|c|c|c|} 
    \hline
    \textbf{Item 1} & \textbf{Item 2} & \textbf{Item 3} & \textbf{Item 4}\\ 
    \hline
    0\% & 100\% & 10\% & 100\% \\
    \hline
    \end{tabular}
\end{center}

\begin{center}
    {Robot's capability \\}
    \begin{tabular}{|c|c|c|c|} 
    \hline
    \textbf{Item 1} & \textbf{Item 2} & \textbf{Item 3} & \textbf{Item 4}\\
    \hline
    100\% & 0\% & 100\% & 10\% \\
    \hline
    \end{tabular}
\end{center}
\label{tab:setup2.4}
\end{table}

\begin{table}[ht]
\small
\caption{Shopping lists (top), human capability in success rate (middle) and robot capability in success rate (bottom) used for simulation experiment Setup 2 with 5 shopping item types.}
\begin{center}
    {Shopping lists \\}
    \begin{tabular}{|c|c|c|c|c|} 
    \hline
    \textbf{Item 1} & \textbf{Item 2} & \textbf{Item 3}  & \textbf{Item 4} & \textbf{Item 5}\\ 
    \hline
    2 & 3 & 2 & 3 & 3 \\
    \hline
    5 & 4 & 2 & 0 & 2 \\
    \hline
    \end{tabular}
\end{center}

\begin{center}
    {Human's capability \\}
    \begin{tabular}{|c|c|c|c|c|} 
    \hline
    \textbf{Item 1} & \textbf{Item 2} & \textbf{Item 3}  & \textbf{Item 4} & \textbf{Item 5}\\
    \hline
    0\% & 100\% & 10\% & 0\% & 100\% \\
    \hline
    \end{tabular}
\end{center}

\begin{center}
    {Robot's capability \\}
    \begin{tabular}{|c|c|c|c|c|} 
    \hline
    \textbf{Item 1} & \textbf{Item 2} & \textbf{Item 3}  & \textbf{Item 4} & \textbf{Item 5}\\
    \hline
    100\% & 0\% & 100\% & 100\% & 10\% \\
    \hline
    \end{tabular}
\end{center}
\label{tab:setup2.5}
\end{table}

\mypara{Setup 3: Varying Number of Shopping Lists}
In this setup, the number of shopping items was fixed at 5 and number of search samples was fixed at 50000. The number of shopping lists was varied from 5 to 10. For an n-shopping list setup, we used the first n shopping lists used for Setup 1. We also used the same actual capability setup as Setup 1.

\subsection{Additional Details for Human Subject Experiments}
In human subject experiments, the shopping lists and the actual capability setup are shown in Table \ref{tab:setup_human}. The TICC-MCP robot initially assumes that 1) the human has perfect capability 2) the human thinks that the robot has perfect capability.

\begin{table}[ht]
\small
\caption{Shopping lists (top), human capability in success rate (middle) and robot capability in success rate (bottom) used for experiments with human subjects.}
\begin{center}
    {Shopping lists \\}
    \begin{tabular}{|c|c|c|c|} 
    \hline
    \textbf{Yellow Cup} & \textbf{Sweet} & \textbf{Cleaning Agent} & \textbf{Green Cup}\\ 
    \hline
    4 & 0 & 0 & 5 \\ 
    \hline
    1 & 1 & 6 & 1 \\
    \hline
    1 & 3 & 5 & 0 \\
    \hline
    3 & 2 & 2 & 1 \\
    \hline
    3 & 1 & 3 & 1 \\
    \hline
    1 & 3 & 3 & 1 \\
    \hline
    7 & 1 & 0 & 0 \\
    \hline
    2 & 3 & 3 & 0 \\
    \hline
    \end{tabular}
\end{center}

\begin{center}
    {Human's capability \\}
    \begin{tabular}{|c|c|c|c|} 
    \hline
    \textbf{Yellow Cup} & \textbf{Sweet} & \textbf{Cleaning Agent} & \textbf{Green Cup}\\  
    \hline
    50\% & 100\% & 0\% & 100\% \\
    \hline
    \end{tabular}
\end{center}

\begin{center}
    {Robot's capability \\}
    \begin{tabular}{|c|c|c|c|} 
    \hline
    \textbf{Yellow Cup} & \textbf{Sweet} & \textbf{Cleaning Agent} & \textbf{Green Cup}\\ 
    \hline
    80\% & 0\% & 100\% & 100\% \\
    \hline
    \end{tabular}
\end{center}
\label{tab:setup_human}
\end{table}

\end{document}